\documentclass[10pt,twocolumn,letterpaper]{article}

\usepackage{cvpr}
\usepackage{times}
\usepackage{epsfig}
\usepackage{graphicx}
\usepackage{amsmath}
\usepackage{amssymb}
\usepackage{subfigure}
\usepackage{multirow}
\usepackage{color}
\usepackage{tabu}
\usepackage{array}
\usepackage[ruled,vlined]{algorithm2e}
\usepackage{bm}

\usepackage[breaklinks=true,bookmarks=false]{hyperref}

\newcommand{\red}[1]{\textcolor[rgb]{1, 0, 0}{#1}}
\newcommand{\blue}[1]{\textcolor[rgb]{0, 0, 1}{#1}}

\cvprfinalcopy 

\def\cvprPaperID{****} 

\begin{document}

\title{Towards Effective Deep Embedding for Zero-Shot Learning}

\author{Lei Zhang$^{1}$, Peng Wang$^{1}$, Lingqiao Liu$^{1}$, Chunhua Shen$^{1}$, Wei Wei$^{2}$, Yannning Zhang$^{2}$, Anton Van Den Hengel$^{1}$\\
$^1$School of Computer Science, The University of Adelaide, Adelaide, 5005, Australia\\
$^2$School of Computer Science, Northwestern Polytechnical University, Xi’an, 710072, China\\
{\tt\small \{lei.Zhang, peng.wang, lingqiao.liu\}@adelaide.edu.au}
}

\maketitle

\begin{abstract}
Zero-shot learning (ZSL) can be formulated as a cross-domain matching problem: after being projected into a joint embedding space, a visual sample will match against all candidate class-level semantic descriptions and be assigned to the nearest class. In this process, the embedding space underpins the success of such matching and is crucial for ZSL. In this paper, we conduct an in-depth study on the construction of embedding space for ZSL and posit that an ideal embedding space should satisfy two criteria: intra-class compactness and inter-class separability. While the former encourages the embeddings of visual samples of one class to distribute tightly close to the semantic description embedding of this class, the latter requires embeddings from different classes to be well separated from each other. Towards this goal, we present a simple but effective two-branch network to simultaneously map semantic descriptions and visual samples into a joint space, on which visual embeddings are forced to regress to their class-level semantic embeddings and the embeddings crossing classes are required to be distinguishable by a trainable classifier. Furthermore, we extend our method to a transductive setting to better handle the model bias problem in ZSL (i.e., samples from unseen classes tend to be categorized into seen classes) with minimal extra supervision. Specifically, we propose a pseudo labeling strategy to progressively incorporate the testing samples into the training process and thus balance the model between seen and unseen classes. Experimental results on five standard ZSL datasets show the superior performance of the proposed method and its transductive extension.
\end{abstract}

\section{Introduction}
With the profit from deep learning~\cite{lecun2015deep}, object recognition~\cite{liang2015recurrent,he2016deep} has gained great success in recent years. The premise of such success is that sufficient annotated samples for each considered object are available for supervised learning~\cite{lecun2015deep,liang2015recurrent}. However, this is often difficult to comply with in real applications due to the prohibitive annotation cost or some harsh conditions for sample collection (\eg, samples in danger scene, newly emerging or identified)~\cite{vinyals2016matching,Sung_2018_CVPR}. Zero-short learning (ZSL)~\cite{frome2013devise,romera2015embarrassingly,qiao2016less,changpinyo2016synthesized,zhang2015zero,kodirov2017semantic,zhang2017learning,
annadani2018preserving,Sung_2018_CVPR} is a task proposed to address an extreme problem, where no annotated samples but only a semantic description are available for a class.

\begin{table*}\small
\renewcommand{\arraystretch}{1.2}
\begin{center}
\begin{tabu} to 1\textwidth {X[l,0.6]|X[c,0.7]|X[c,0.35]|X[c,0.35]|X[l]}
\hline
Method & Type of embedding space & Intra-class compactness & Inter-class separability & \multicolumn{1}{c}{Comments}\\
\hline
DAP, IAP~\cite{lampert2014attribute}, CMT~\cite{socher2013zero}, DEVISE~\cite{frome2013devise} & {Semantic space} & {$\surd$} & {$\times$} & {1) Fixing semantic space and assuming it has sufficient discriminability; 2) Bias toward seen classes~\cite{chao2016empirical}}\\
\hline
PSR~\cite{annadani2018preserving}, DEM~\cite{zhang2017learning} & {Visual space} & {$\surd$} & {$\times$} & {1) Fixing visual space and assuming it has sufficient discriminability; 2) Bias toward seen classes~\cite{chao2016empirical}}\\
\hline
CONSE~\cite{norouzi2013zero}, SSE~\cite{zhang2015zero}, LATEM~\cite{xian2016latent}, ALE~\cite{akata2016label}, SJE~\cite{akata2015evaluation}, ESZSL~\cite{romera2015embarrassingly}, SYNC~\cite{changpinyo2016synthesized}, SAE~\cite{kodirov2017semantic}, GFZSL~\cite{verma2017simple}, RN~\cite{Sung_2018_CVPR}  & {Latent intermediate space} & {$\times$} & {$\surd$} & {1) Implicitly learning an intermediate space by fitting a compatibility function; 2) Bias toward seen classes~\cite{chao2016empirical}}\\
\hline
Ours & Latent intermediate space & $\surd$ & $\surd$ & {1) Explicitly learning an intermediate space; 2) Less effected by bias problem}\\
\hline
\end{tabu}
\end{center}
\caption{Different embedding spaces learned in existing ZSL methods.}
\label{table:property}
\vspace{-0.3cm}
\end{table*}

\begin{figure*}
\setlength{\abovecaptionskip}{0pt}
\begin{center}
\subfigure[Visual space (all classes)]{\includegraphics[height=1.9in,width=2.2in,angle=0]{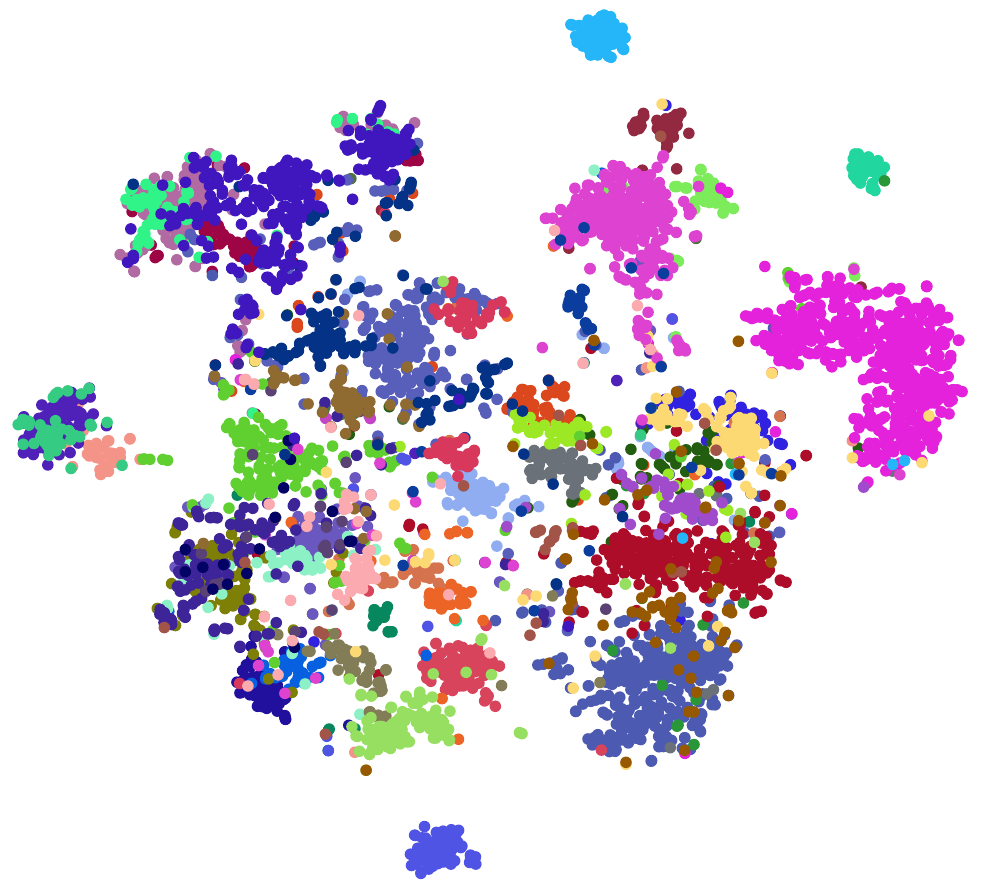}}
\subfigure[Ours (all classes)]{\includegraphics[height=1.9in,width=2.2in,angle=0]{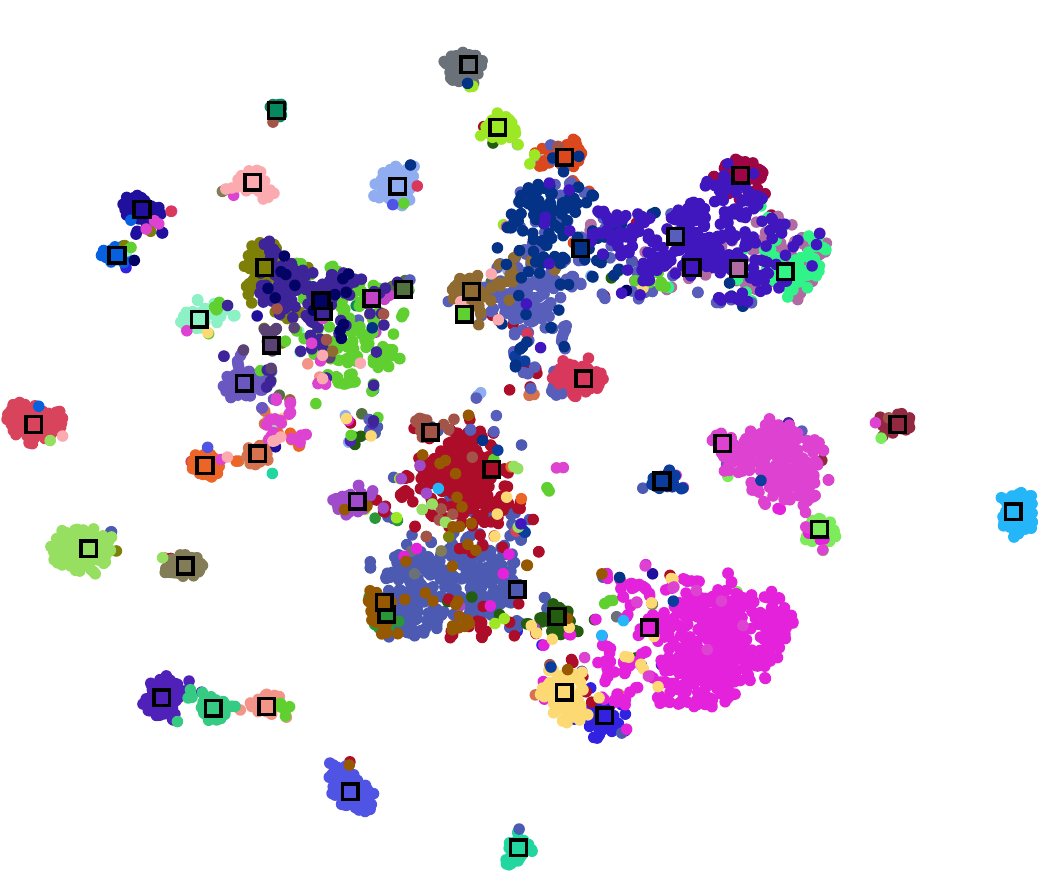}}
\subfigure[Ours\_ex (all classes)]{\includegraphics[height=1.9in,width=2.2in,angle=0]{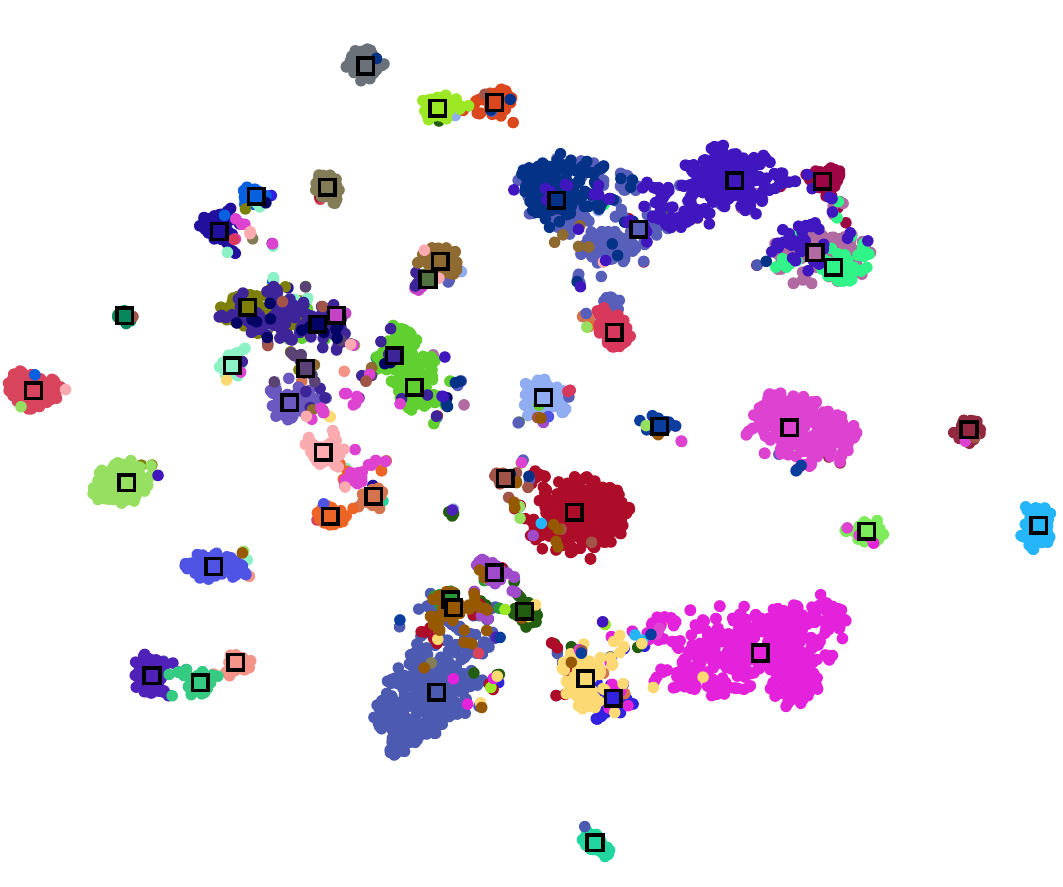}}
\\
\vspace{-0.3cm}
\subfigure[Visual space (unseen classes)]{\includegraphics[height=1.9in,width=2.2in,angle=0]{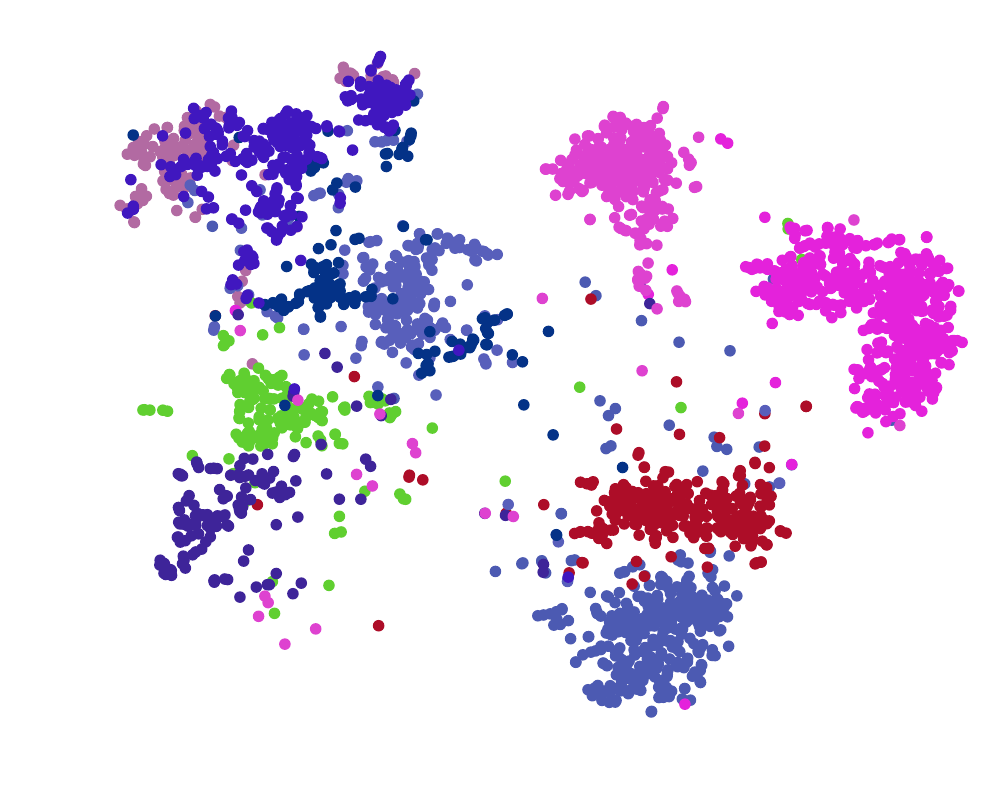}}
\subfigure[Ours (unseen classes)]{\includegraphics[height=1.9in,width=2.2in,angle=0]{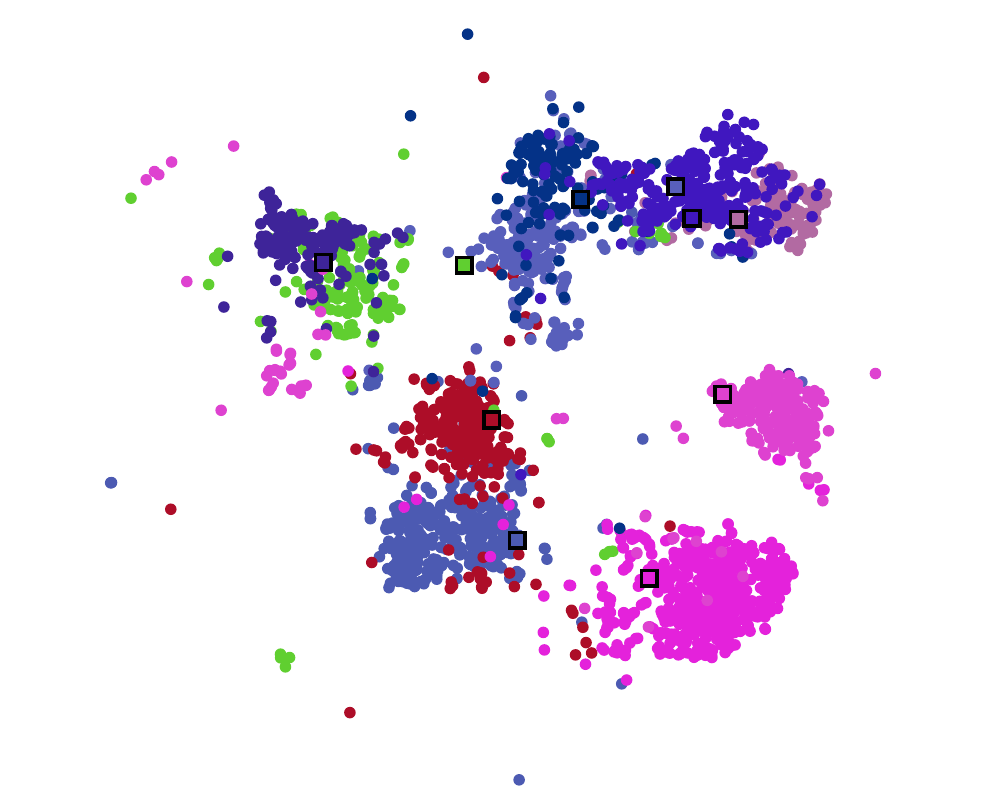}}
\subfigure[Ours\_ex (unseen classes)]{\includegraphics[height=1.9in,width=2.2in,angle=0]{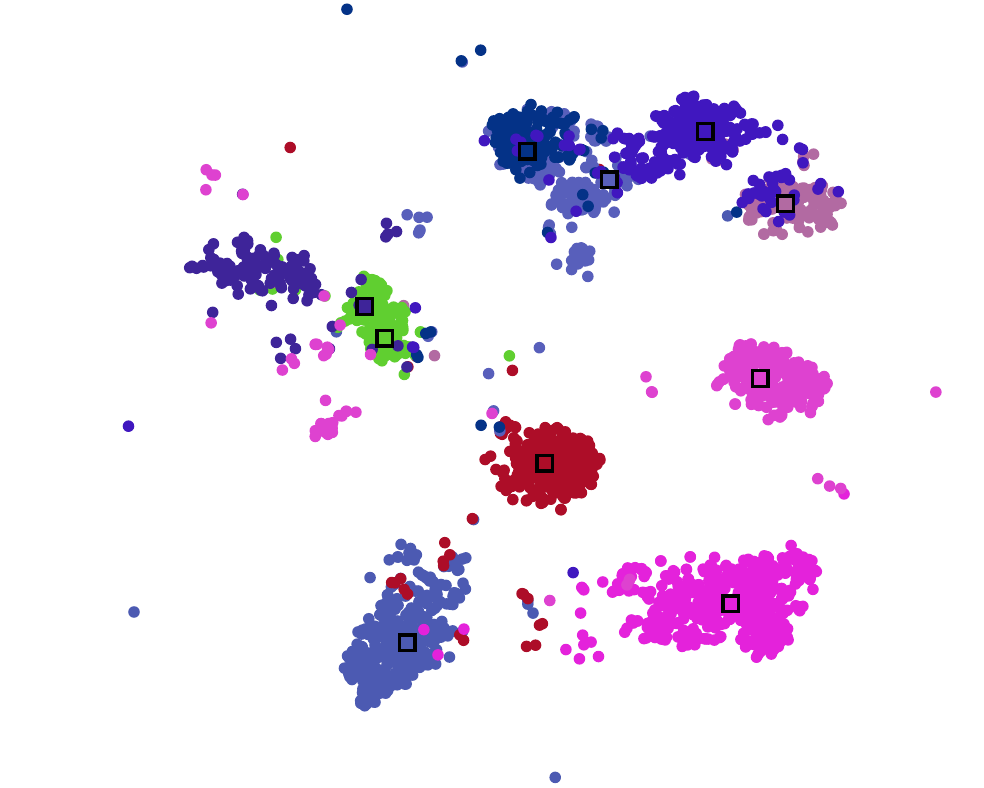}}
\end{center}
\vspace{-0.3cm}
\caption{Visualization of the distribution of samples from the AwA1 dataset~\cite{lampert2014attribute} in different embedding spaces, including the visual space~\cite{zhang2017learning} and those learned in our method (\ie, '{\texttt{Ours}}' and '{\texttt{Ours\_ex}}' denote our method and its extension to transductive setting). Dots and squares represent the embeddings of visual samples and the class-level semantic descriptions, respectively. Different colors represent different classes. The first row shows the distribution of embeddings from both seen and unseen classes, while the second row shows the distribution of embeddings from unseen classes.}
\vspace{-0.4cm}
\label{fig:embedding}
\end{figure*}

In contrast to conventional supervised learning, ZSL attempts to recognize samples from {\textit{unseen classes}} through exploiting the semantic connections between these classes and some other {\textit{seen classes}} which have sufficient annotated training samples. In ZSL, each class is represented by a semantic vector (\eg, attributes~\cite{farhadi2009describing}, word vector~\cite{socher2013zero,frome2013devise}, or even an encoding vector for a sentence~\cite{socher2013zero}), which composes the {\textit{semantic space}} shared by both seen and unseen classes, and the visual representations of the visual samples constitute the {\textit{visual space}}.

Essentially, ZSL can be formulated as a cross-domain matching problem: after being projected into a joint embedding space, a visual sample will match against all candidate class-level semantic descriptions and be assigned to the nearest class. In this process, the embedding space underpins the success of matching and stimulates research on approaches to learn an effective embedding space. Generally, the embedding schemes can be categorized into two groups based on learning unidirectional mapping functions or bidirectional mapping functions. The former fixes either \textit{semantic space}~\cite{socher2013zero,frome2013devise, lampert2014attribute,fu2016semi} or \textit{visual space}~\cite{zhang2017learning,shigeto2015ridge,annadani2018preserving} as the anchor space and learn a mapping function to align the other modality onto this space. These approaches are based on a common assumption that the chosen anchor space has sufficient discriminability to distinguish different classes either seen or unseen. But in practice, this assumption may not hold for state-of-the-art semantic representations or even for powerful deep features \cite{zhang2017learning} trained on large-scale external dataset, as shown in Figure~\ref{fig:embedding} (a)(d). Another line of works project both visual samples and the class-level semantic descriptions into a latent intermediate space via fitting a compatibility function between two modalities in order that the visual samples can be successfully distinguished~\cite{romera2015embarrassingly,akata2015evaluation,lei2015predicting,
zhang2016zero,Sung_2018_CVPR}. These approaches, however, suffer from a common drawback that they ignore the intra-class compactness and the resultant large intra-class variance thereon can hinder the generalization capacity in ZSL.

In this paper, we posit that an ideal embedding space should satisfy two criteria: intra-class compactness and inter-class separability, which promote the generalization capacity in ZSL in different ways. While the former encourages the visual embeddings of a class to distribute tightly close to the semantic embedding of this class, the latter forces the embeddings crossing classes to be distinguishable. Towards this goal, we design a simple but effective two-branch network to simultaneously map the semantic descriptions and visual representations into a joint space. We design a new loss function which is composed of two terms: a regression term and a classification term. The regression term minimizes the absolute distance between the embeddings of a visual sample and its class-level semantic description. The classification term forces the embeddings crossing classes can be distinguished by learning an auxiliary classifier. Through learning the embedding space this way, a visual sample will be close to its class-level semantic identification and far away from identifications of other classes, thus enhancing the generalization capacity in ZSL, as shown in Figure~\ref{fig:embedding} (b)(e).

Furthermore, we also extend our method to a transductive setting to better handle the model bias problem~\cite{chao2016empirical} in ZSL: a sample from unseen class has high probability to be assigned to a seen class. Specifically, we propose a pseudo labeling strategy to progressively incorporate the testing samples with pseudo labels into the training process, thus enabling the learned embeddings to better balance  between seen and unseen classes, as shown in Figure~\ref{fig:embedding} (c)(f). Although our method employs the testing data, we require no other extra supervision information. A summary of existing embedding schemes and our embedding scheme for ZSL can be found in Table~\ref{table:property}.

Experimental results on five standard ZSL datasets show the superior performance of the proposed method as well as its transductive extension. 

\begin{figure*}
\setlength{\abovecaptionskip}{0pt}
\begin{center}
\includegraphics[height=1.3in,width=6.3in,angle=0]{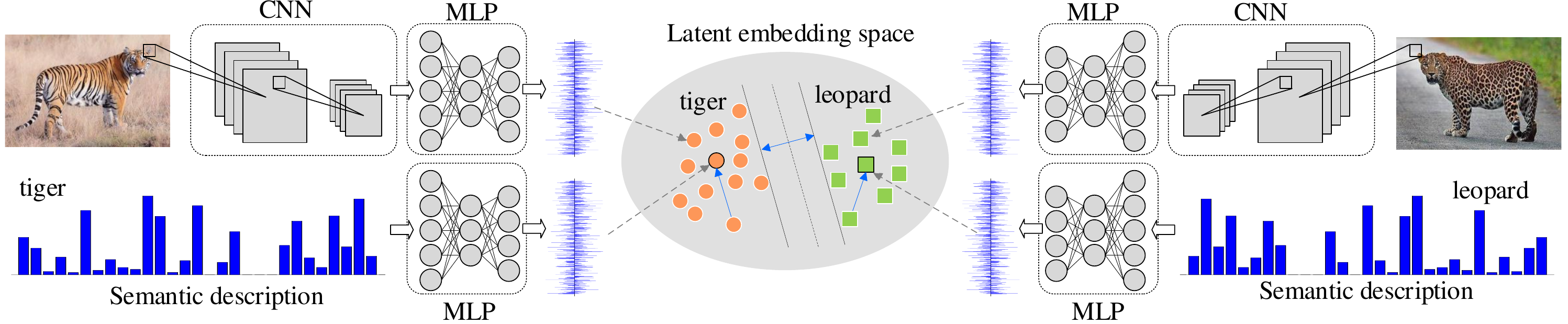}
\end{center}
\vspace{-0.3cm}
\caption{The architecture of the proposed two-branch deep embedding model.}
\vspace{-0.1cm}
\label{fig:flow}
\end{figure*}

\section{Related works}
\paragraph{Embedding model} Based on whether unidirectional mapping function or bidirectional mapping functions are learned, existing ZSL methods can be roughly divided into two groups. While the former fixes either the semantic space or visual space as an anchor space and aligns the other modality onto this anchor space, the latter simultaneously encodes these two spaces in order that they can match each other. For simplicity, we call these three types of approaches: semantic space, visual space, and latent intermediate space. 1) {\textit{Semantic space}}. This group of methods exploit the semantic space as the embedding space~\cite{socher2013zero,frome2013devise, lampert2014attribute,fu2016semi}. For example, Socher et al.~\cite{socher2013zero} and Frome et al.~\cite{frome2013devise} learn to project the pre-computed shallow features or deep features of visual samples to the semantic descriptions and adopt a nearest neighbor (NN) classifier to assign visual samples to their classes. However, due to the limited representation capacity of a low-dimensional semantic representation, \eg visual attributes \cite{farhadi2009describing}, word vector \cite{socher2013zero}, these approaches may suffer from the hubness problem~\cite{zhang2017learning}: a semantic description can be nearest neighbour to visual samples from multiple classes. 2) {\textit{Visual space}}. This line of research learns to align the semantic description of a class onto the fixed visual space of this class~\cite{zhang2017learning,shigeto2015ridge}. They attempt to mitigate the hubness problem through exploiting the discriminative capacity of visual features, such as deep features trained on large-scale external dataset~\cite{zhang2017learning}. Recently, Annadani \etal~\cite{annadani2018preserving} take a step further by explicitly modeling the inter-class semantic relationships (\eg, similar or dissimilar) when learning the embedding space. However, visual space often shows large intra-class variance which means samples from different classes may be difficult to be dispersed from each other by an obvious margin, as shown in Figure~\ref{fig:embedding} (a)(d). 3) {\textit{Latent intermediate space}}. This class of methods project both visual and semantic features into a latent intermediate space and fit a compatibility function to distinguish samples from different classes~\cite{romera2015embarrassingly,akata2015evaluation,lei2015predicting,
zhang2016zero,Sung_2018_CVPR}. For example, recently, Sung \etal~\cite{Sung_2018_CVPR} employ the relation network~\cite{santoro2017simple} as the compatibility function and learn this compatibility function as well as two separate mapping functions in an end-to-end manner. A common problem for this line of works is that they focus more on distinguishing the training samples but ignoring limiting the intra-class variance within each class, which can hinder the generalization capacity in ZSL.

\paragraph{Inductive \& transductive ZSL} According to whether the testing data is used for training or not, existing ZSL methods can be grouped into two categories. 1) {\textit{Inductive ZSL}} methods~\cite{frome2013devise,changpinyo2016synthesized,zhang2015zero,kodirov2017semantic,zhang2017learning,
annadani2018preserving,Sung_2018_CVPR}. They only utilize the labeled training samples for model training. 2) {\textit{Transductive ZSL}} methods~\cite{kodirov2015unsupervised,song2018transductive}. They also introduce unlabeled testing samples into model training to handle the model bias problem~\cite{chao2016empirical} in ZSL. For example, on top of the seen classes, Song \etal~\cite{song2018transductive} learn another super-category to represent all the unseen classes. By encouraging the samples from unseen classes to be classified into this super-category, they can alleviate the model bias problem. But their method is based on a strong assumption, that is they need to pre-know a testing sample is from a seen or unseen class. This supervision information, however, cannot be obtained in practical generalized ZSL setting, where the testing data is the combination of samples from both seen and unseen classes, as shown in Table~\ref{table:conv_gen}. Our method exploits the testing samples for model training by automatically generating the pseudo labels to them, which requires no extra supervision information. In this sense, our method is more practical and more appropriate to cope with generalized ZSL setting.

\begin{table}\scriptsize%
\renewcommand{\arraystretch}{1.3}
\begin{center}
\begin{tabu} to 0.5\textwidth {X[l,0.4]|X[c,0.8]|X[c]}
\hline
ZSL Setting & Training (labeled) data & Testing (unlabeled) data \\
\hline
Conventional & Samples from seen classes & Samples from unseen classes\\
\hline
Generalized & Samples from seen classes & Samples from seen classes and unseen classes\\
\hline
\end{tabu}
\end{center}
\caption{Conventional ZSL setting vs. generalized ZSL setting.}
\label{table:conv_gen}
\vspace{-0.3cm}
\end{table}

\section{Methodology}
Provided that a training set with $N$ samples is given as $\mathcal{D}_{tr}=\{(\mathbf{x}_i, y^s_i), i=1,...,N\}$, where $\mathbf{x}_i$ denotes the $i$-th visual sample (\eg, image) with class label $y^s_i\in{\mathcal{Y}^{tr}}$ and $\mathcal{Y}^{tr}$ is the label set of all seen classes. In the testing phase, ZSL aims at predicting the label $y^t_j\in{\mathcal{Y}^{ts}}$ for a new sample $\mathbf{x}_j$. $\mathcal{Y}^{ts}$ denotes the label set of all unseen classes and $\mathcal{Y}^{tr} \cap \mathcal{Y}^{ts} = \varnothing$. Each class of $y^s_i$ or $y^t_j$ is associated with a semantic description (\eg, attribute vector) $\mathbf{z}^s_{i}$ or $\mathbf{z}^t_{j}$. For the generalized ZSL setting~\cite{xian2017zero}, the only difference is that the test sample $\mathbf{x}_j$ may come from either unseen classes or seen classes, \ie, $y^t_j\in{\mathcal{Y}^{tr} \cup \mathcal{Y}^{ts}}$.

\subsection{Proposed deep embedding model}
In this study, we attempt to learn an effective intermediate embedding space which shows both intra-class compactness and inter-class separability. To this end, we design a two-branch deep embedding model to simultaneously embed the visual and semantic representations. As illustrated in Figure~\ref{fig:flow}, the network is composed of two branches: a visual embedding branch and a semantic embedding branch.
For the visual embedding branch, an image is firstly fed into a pre-trained deep convolutional network to obtain its visual representation $\mathbf{x}_i$\footnote{We employ the same notation $\mathbf{x}_i$ to denote both the image and its visual representation to avoid the abuse of notations.}. Then, a multilayer perceptron (MLP) ${\mathbf{\theta}_v}(\mathbf{x}_i)$ is learned to embed the visual representation into a latent space. Parallelly, for the semantic embedding branch, another MLP ${\psi_{\mathbf{\theta}_s}(\mathbf{z}^s_i)}$ is applied on the semantic description $\mathbf{z}^s_i$ to map the semantic description into the same latent space. On top of the intermediate latent space, we define a novel loss function which is composed of two terms: a regression term and a classification term. While the regression term minimizes the distance between the embedding of a visual sample and its class-level semantic embedding, the classification term distinguishes the embeddings crossing categories by learning an auxiliary classifier. Mathematically, this process can be formulated as,
\begin{equation}\label{eq:eq1}
\begin{aligned}
\min\limits_{\mathbf{\theta}_v, \mathbf{\theta}_s, \mathbf{W}} &\sum\limits^N_{i=1}\|\phi_{\mathbf{\theta}_v}(\mathbf{x}_i) - \psi_{\mathbf{\theta}_s}(\mathbf{z}^s_i)\|^2_2 + \lambda\mathcal{L}(\mathbf{W}^T\phi_{\mathbf{\theta}_v}(\mathbf{x}_i), y^s_i)\\
+ &\eta\left(\|\mathbf{\theta}_v\|^2_2 + \|\mathbf{\theta}_s\|^2_2 + \|\mathbf{W}\|^2_2\right),
\end{aligned}
\end{equation}
where ${\mathbf{\theta}_v}$ and ${\mathbf{\theta}_s}$ are the parameters involved in the MLPs, $\mathbf{W}$ denotes the linear classifier to be learned, and $\mathcal{L}(\cdot,\cdot)$ is a classification loss, which is chosen to be CrossEntropy in this paper. Also, to prevent overfitting, we constrain the $\ell_2$ norm of all parameters, which are weighted by $\eta$. Given the learned parameter ${\mathbf{\theta}_v}$ and ${\mathbf{\theta}_s}$, the label for a testing sample $\mathbf{x}^t_j$ can be predicted as,
\begin{equation}\label{eq:eq2}
\begin{aligned}
\hat{y}^t_j=\min\limits_{y\in{\mathcal{Y}^{ts}}} \|\phi_{\mathbf{\theta}_v}(\mathbf{x}^t_j) - \psi_{\mathbf{\theta}_s}(\mathbf{z})\|^2_2
\end{aligned}
\end{equation}
where $\mathbf{z}$ denotes the semantic description associated with the label $y$. For the generalized ZSL setting, we only need to modify the solution space of label as $y\in{\mathcal{Y}^{tr}\cup \mathcal{Y}^{ts}}$.

Note that two-branch embedding structure has been employed in some previous works~\cite{lei2015predicting,akata2015evaluation} to learn an intermediate latent embedding space. These methods tend to model the latent embedding space implicitly via learning a compatibility function between visual samples and their associated class-level semantic descriptions to distinguish samples from different classes. In this study, we embrace an explicit way to learn such an intermediate embedding space. On one hand,  we explicitly project the samples from two modalities (\ie, semantic and visual modalities) into a joint space and minimize their distances thereon to increase the intra-class compactness, viz., reducing the intra-class variance. On the other hand, an auxiliary linear classifier is learned to force the samples from different classes to be well separated on the embedding space, viz., increasing the inter-class separability. The classifier can also prevent a trivial solution that may be caused by minimizing the distances crossing modalities \footnote{By minimizing the Euclidean distance, both modalities tend to be mapped to $0$.}. Both of theses two aspects turn out to be important in maintaining the generalization capacity in ZSL.

\subsection{Optimization}
Considering that $\mathbf{\theta}_v$ is coupled with $\mathbf{\theta}_s$ in Eq.~\eqref{eq:eq1}, we adopt the alternative minimizing scheme~\cite{zhang2018unsupervised} to reduce the optimization problem in Eq.~\eqref{eq:eq1} into two subproblems and then optimize each of them alternatively until convergence. In this study, we term these two subproblems, the visual embedding problem and the semantic embedding problem.

{\textbf{Visual embedding problem}}. Given $\mathbf{\theta}^{(t)}_s$ in the $t$-th iteration, we can estimate $\mathbf{\theta}^{(t+1)}_v$ and $\mathbf{W}^{(t+1)}$ by solving the following problem
\begin{equation}\label{eq:eq4}
\begin{aligned}
\min\limits_{\mathbf{\theta}_v, \mathbf{W}} &\sum\limits^N_{i=1}\|\phi_{\mathbf{\theta}_v}(\mathbf{x}_i) - \psi_{{\mathbf{\theta}^{(t)}_s}}(\mathbf{z}^s_i)\|^2_2 + \lambda\mathcal{L}(\mathbf{W}^T\phi_{\mathbf{\theta}_v}(\mathbf{x}_i), y^s_i)\\
+&\eta\left(\|\mathbf{\theta}_v\|^2_2 + \|\mathbf{W}\|^2_2\right).
\end{aligned}
\end{equation}

{\textbf{Semantic embedding problem}}. Given $\mathbf{\theta}^{(t+1)}_v$, the subproblem for estimating $\mathbf{\theta}^{(t+1)}_s$ can be formulated as
\begin{equation}\label{eq:eq5}
\begin{aligned}
\min\limits_{\mathbf{\theta}_s} \sum\limits^N_{i=1}\|\phi_{\mathbf{\theta}^{(t+1)}_v}(\mathbf{x}_i) - \psi_{\mathbf{\theta}_s}(\mathbf{z}^s_i)\|^2_2 + \eta\|\mathbf{\theta}_s\|^2_2,
\end{aligned}
\end{equation}

Since both subproblems above are differential, we adopt back-propagation algorithm~\cite{he2016deep} to solve them. The training pipeline of our model is summarized into Algorithm~\ref{alg:Algorithm}.

\begin{algorithm}
\caption{Learning embedding space}
\label{alg:Algorithm}
\KwIn{Training set $\mathcal{D}_{tr}$, class-level semantic descriptions, scalar $\lambda$ and $\eta$;}
\textbf{Loop}: for $t=1,2,...,T$ \textbf{do}\\
${\kern 8pt}(a)$ Solve visual embedding problem as Eq.~\eqref{eq:eq4};\\
${\kern 8pt}(b)$ Solve semantic embedding problem as Eq.~\eqref{eq:eq5};\\
\textbf{End for}\\
\KwOut{Model parameters $\mathbf{\theta}_v$ and $\mathbf{\theta}_s$.}
\end{algorithm}

\begin{algorithm}
\caption{Learning embedding space with pseudo labeling strategy}
\label{alg:Algorithm2}
\KwIn{Training set $\mathcal{D}_{tr}$, class-level semantic descriptions, testing images $\{\mathbf{x}^t_j\}$, scalar $\lambda$, $\eta$, $M=M_0$;}
\textbf{Loop}: for $r=1,2,...,R$ \textbf{do}\\
${\kern 4pt}1.$ Learning embedding space as Algorithm~\ref{alg:Algorithm};\\
${\kern 4pt}2.$ Training dataset augmentation:\\
${\kern 8pt}(a)$ Predict $\{\hat{y}^t_j\}$ for $\{\mathbf{x}^t_j\}$ as Eq.~\eqref{eq:eq2};\\
${\kern 8pt}(b)$ Select the top-$M$ high-confidence samples from\\
${\kern 20pt}$ each unseen class;\\
${\kern 8pt}(c)$ Augment $\mathcal{D}_{tr}$ with selected samples;\\
${\kern 8pt}(d)$ Update $M = M_0 * (r + 1)$;\\
\textbf{End for}\\
\KwOut{Model parameters $\mathbf{\theta}_v$ and $\mathbf{\theta}_s$.}
\end{algorithm}

\section{Pseudo labeling strategy}
In most ZSL methods, only training samples from seen classes are utilized to learn the embedding space. Thus, the learned embedding space can produce bias towards seen classes~\cite{chao2016empirical}. To better cope the bias problem, we present a pseudo labeling strategy to extend our method to the transductive setting~\cite{kodirov2015unsupervised,song2018transductive} where unlabeled testing samples also can be exploited for model training.

Specifically, given a set of testing samples (which may come from both seen and unseen classes in generalized ZSL setting), we first predict their pseudo labels with the learned embedding space as Eq.~\eqref{eq:eq2}. Then, according to the generated pseudo labels as well as the visual-semantic gap (\ie, $\|\phi_{\mathbf{\theta}_v}(\mathbf{x}^t_j) - \psi_{\mathbf{\theta}_s}(\mathbf{z})\|^2_2$), we select the top-$M$ high-confidence testing samples that are predicted to belong to an unseen class and incorporate them as well as their predicted pseudo labels into the training set $\mathcal{D}_{tr}$. With the augmented $\mathcal{D}_{tr}$, the optimization objective becomes,
\begin{equation}\label{eq:eq3}
\begin{aligned}
\min\limits_{\mathbf{\theta}_v, \mathbf{\theta}_s, \mathbf{W}} &\sum\limits^N_{i=1}\|\phi_{\mathbf{\theta}_v}(\mathbf{x}_i) - \psi_{\mathbf{\theta}_s}(\mathbf{z}^s_i)\|^2_2 + \lambda\mathcal{L}(\mathbf{W}^T\phi_{\mathbf{\theta}_v}(\mathbf{x}_i), y^s_i)\\
+&\sum\limits^{CM}_{i=1}\|\phi_{\mathbf{\theta}_v}({\tilde{\mathbf{x}}}_i) - \psi_{\mathbf{\theta}_s}({\tilde{\mathbf{z}}}^s_i)\|^2_2 + \lambda\mathcal{L}(\mathbf{W}^T\phi_{\mathbf{\theta}_v}({\tilde{\mathbf{x}}}_i), {\tilde{y}}^s_i)\\
+&\eta\left(\|\mathbf{\theta}_v\|^2_2 + \|\mathbf{\theta}_s\|^2_2 + \|\mathbf{W}\|^2_2\right),
\end{aligned}
\end{equation}
where $C$ denotes the number of unseen classes and ${\tilde{\mathbf{x}}}_i$ is the $i$-th selected sample. ${\tilde{y}}^s_i$ and ${\tilde{\mathbf{z}}}^s_i$ are the corresponding pseudo label and the class-level semantic description.

In Eq.~\eqref{eq:eq3}, the regression loss, \ie, $\|\phi_{\mathbf{\theta}_v}({\tilde{\mathbf{x}}}_i) - \psi_{\mathbf{\theta}_s}({\tilde{\mathbf{z}}}^s_i)\|^2_2$, forces each testing visual sample ${\tilde{\mathbf{x}}}_i$ from a predicted unseen class to coincide with its semantic description ${\tilde{\mathbf{z}}}^s_i$. The classification loss, \ie, $\mathcal{L}(\mathbf{W}^T\phi_{\mathbf{\theta}_v}({\tilde{\mathbf{x}}}_i), {\tilde{y}}^s_i)$, forces the separability between any two classes, including seen and unseen classes, in the augmented training data. Thus, the proposed strategy above is able to well alleviate the bias problem, and consequently lead to better ZSL performance. Resorting to a coarse-to-fine strategy, we repeat this pseudo-labelling process to incorporate increasing number of testing samples into training and progressively calibrate the embedding space. To minimize the risk of introducing incorrectly labeled testing samples, we set $M$ to be a small value at beginning and then gradually increase $M$ to select more samples. The pipeline of this strategy is summarized in Algorithm~\ref{alg:Algorithm2}. Note that although the proposed strategy utilizes the testing samples in the training process, it requires no other extra supervision. For example, we do not need to know whether a testing sample is from a seen or unseen class in advance~\cite{song2018transductive}.

In Figure~\ref{fig:embedding} (c)(f), we visualize the embedding space learned by our method with the pseudo labeling strategy in Eq.~\eqref{eq:eq3} with 9 rounds of calibration. It can be seen that the embeddings of visual samples from unseen classes stay close to their associated class-level semantic embeddings and far away from other semantic embeddings, especially those of seen classes. 

\section{Experiment}\label{sec:experiments}

\paragraph{Datasets} We evaluate the proposed method on five standard ZSL datasets. {\textbf{AwA1}}~\cite{lampert2014attribute} contains $30,745$ images of $50$ classes of animals where $40$ classes are seen for training while the remaining $10$ classes are unseen during training. Each class is associated with a $85$-dimension continuous attribute vector. {\textbf{AwA2}}~\cite{xian2017zero} consists of $37,322$ images of the same $50$ classes but with images not overlapped with {\textbf{AwA1}}. {\textbf{CUB}} (Caltech-UCSD Birds-200-2011)~\cite{wah2011multiclass} contains $11,788$ images of $200$ fine-grained bird species. A standard split divides these bird species into $150$ seen classes and $50$ unseen classes. For each class, a $312$-dimension continuous attribute vector is provided. {\textbf{SUN}}~\cite{patterson2012sun} consists of $14,340$ images to describe $717$ scene categories where $645$ classes are selected for training and the remaining $72$ classes are used for testing. A $102$-dimension continuous attribute vector is provided for each class as semantic description. {\textbf{aPY}}~\cite{farhadi2009describing} is a small-scale dataset with 64 attributes. It contains $32$ classes, where $20$ Pascal classes are used for training and $12$ Yahoo classes are used for testing.
\vspace{-0.4cm}

\begin{table}\scriptsize%
\renewcommand{\arraystretch}{1.3}
\begin{center}
\begin{tabular}{l|c}
\hline
Function & Implementation\\
\hline
$\phi_{\theta_v}$ & Fully Connected (FC) layer + Rectified Linear Unit (ReLU) \\
$\psi_{\theta_s}$ & FC{\red{~\footnotemark[3]}} + ReLu + FC + ReLu\\
\hline
\end{tabular}
\end{center}
\caption{Implementation of two non-linear mapping functions.}
\label{table:function}
\vspace{-0.3cm}
\end{table}

\paragraph{ZSL settings} In this study, we conduct experiments under both conventional ZSL setting~\cite{song2018transductive} and generalized ZSL setting~\cite{Sung_2018_CVPR}. In the conventional ZSL, testing samples are restricted to unseen classes, while in the generalized ZSL, they may come from either seen classes or unseen classes, as shown in Table~\ref{table:conv_gen}.
\vspace{-0.3cm}

\paragraph{Implementation details} For fair comparison, following \cite{xian2017zero}, we adopt $2048$-dimensional ResNet101~\cite{he2016deep} feature as image representation in our method (\eg, as well as all comparing methods in this study). We utilize two MLPs $\phi_{\theta_v}$, $\psi_{\theta_s}$ to embed both the visual features and semantic descriptions into a $1024$-dimensional space. The details of these two MLPs~\footnotetext[3]{Since the dimension of input semantic vector differs in five ZSL datasets, we set the output dimension of the first FC in $\psi_{\theta_s}$ as \\ (dimension of the input semantic vector + 1024)/2.} can be found in Table~\ref{table:function}. The scalar $\lambda$ and $\eta$ are determined by cross validation on each benchmark. We train our network with Adam \cite{kingma2014adam} optimizer with learning rate $1e^{-4}$. In Algorithm~\ref{alg:Algorithm} and Algorithm~\ref{alg:Algorithm2}, the maximal iteration of the loop is set as $T=50$ and $R=10$. In Algorithm~\ref{alg:Algorithm2}, the initial number of selected samples is set as $M_0=40$. For simplicity, in the following experiments we denote our method in Eq.~\eqref{eq:eq1} as '{\texttt{Ours}}', while its transductive extension in Eq.~\eqref{eq:eq3} is denoted as '{\texttt{Ours\_ex}}'. Our implementation is based on Pytorch.
\vspace{-0.3cm}

\paragraph{Comparing methods} We compare our method to 17 existing ZSL methods.
Among them, DAP~\cite{lampert2014attribute},
IAP~\cite{lampert2014attribute}, CMT~\cite{socher2013zero} and DEVISE~\cite{frome2013devise} are semantic embedding space based methods. PSR~\cite{annadani2018preserving} and DEM~\cite{zhang2017learning} are visual embedding space based methods. The rest, \eg, CONSE~\cite{norouzi2013zero}, SSE~\cite{zhang2015zero}, LATEM~\cite{xian2016latent}, ALE~\cite{akata2016label}, SJE~\cite{akata2015evaluation}, ESZSL~\cite{romera2015embarrassingly}, SYNC~\cite{changpinyo2016synthesized}, SAE~\cite{kodirov2017semantic}, GFZSL~\cite{verma2017simple}, RN~\cite{Sung_2018_CVPR} and QFSL~\cite{song2018transductive}, are the latent intermediate embedding space based methods. Note that all these comparing methods utilize the same visual features, \eg., $2048$-dimensional ResNet101~\cite{he2016deep} features, as our method for image representation.

\begin{table}\small
\renewcommand{\arraystretch}{1.2}
\begin{center}
\begin{tabular}{l|c|c|c|c|c}
\hline
Method & {\textbf{AwA1}} & {\textbf{AwA2}} & {\textbf{CUB}} & {\textbf{SUN}} & {\textbf{aPY}}\\
\hline
DAP~\cite{lampert2014attribute} & 44.1 & 46.1 & 40.0 & 39.9 & 33.8\\
IAP~\cite{lampert2014attribute} & 35.9 & 35.9 & 24.0 & 19.4 & 36.6\\
CMT~\cite{socher2013zero} & 39.5 & 37.9 & 34.6 & 39.9 & 28.0\\
DEVISE~\cite{frome2013devise} & 54.2 & 59.7 & 52.0 & 56.5 & {\textbf{39.8}}\\
\hline
PSR~\cite{annadani2018preserving} & - & 63.8 & 56.0 & 61.4 & 38.4\\
DEM~\cite{zhang2017learning} & 68.4 & {\blue{67.1}} & 51.7 & 40.3 & 35.0\\
\hline
CONSE~\cite{norouzi2013zero} & 45.6 & 44.5 & 34.3 & 38.8 & 26.9\\
SSE~\cite{zhang2015zero} & 60.1 & 61.0 & 43.9 & 51.5 & 34.0\\
LATEM~\cite{xian2016latent} & 55.1 & 55.8 & 49.3 & 55.3 & 35.2\\
ALE~\cite{akata2016label} & 59.9 & 62.5 & 54.9 & 58.1 & 39.7\\
SJE~\cite{akata2015evaluation} & 65.6 & 61.9 & 53.9 & 53.7 & 32.9\\
ESZSL~\cite{romera2015embarrassingly} & 58.2 & 58.6 & 53.9 & 54.5 & {\blue{38.3}}\\
SYNC~\cite{changpinyo2016synthesized} & 54.0 & 46.6 & 55.6 & 56.3 & 23.9\\
SAE~\cite{kodirov2017semantic} & 53.0 & 54.1 & 33.3 & 40.3 & 8.3\\
GFZSL~\cite{verma2017simple} & 68.3 & 63.8 & 49.3 & 60.6 & 38.4\\
RN~\cite{Sung_2018_CVPR} & 68.2 & 64.2 & 55.6 & - & -\\
\hline
Ours & {\blue{70.1}} & 66.5 & {\blue{57.1}} & {\textbf{62.4}} & 20.4\\
Ours\_ex & {\textbf{85.3}} & {\textbf{77.5}} & {\textbf{67.8}} & {\blue{61.6}} & 31.3\\
\hline
\end{tabular}
\end{center}
\caption{Recognition accuracy on five benchmarks under the conventional ZSL setting. The best results are in blod, and the second best results are in blue. The competing approaches are grouped into three categories according to the type of their embedding spaces (\it from top to bottom: semantic space, visual space and the latent intermediate space).}
\label{table:con_ZSL}
\vspace{-0.3cm}
\end{table}

\begin{table*}\small
\renewcommand{\arraystretch}{1.2}
\begin{center}
\begin{tabular}{l|c|c|c|c|c|c|c|c|c|c|c|c|c|c|c}
\hline
 & \multicolumn{3}{c|}{\textbf{AwA1}} & \multicolumn{3}{c|}{\textbf{AwA2}} & \multicolumn{3}{c|}{\textbf{CUB}} & \multicolumn{3}{c|}{\textbf{SUN}} & \multicolumn{3}{c}{\textbf{aPY}}\\
\hline
Method & {\textbf{ts}} & {\textbf{tr}} & {\textbf{H}} & {\textbf{ts}} & {\textbf{tr}} & {\textbf{H}} & {\textbf{ts}} & {\textbf{tr}} & {\textbf{H}} & {\textbf{ts}} & {\textbf{tr}} & {\textbf{H}} & {\textbf{ts}} & {\textbf{tr}} & {\textbf{H}}\\
\hline
DAP~\cite{lampert2014attribute} & 0.0 & 88.7 & 0.0 & 0.0 & 84.7 & 0.0 & 1.7 &  67.9 & 3.3 & 4.2 & 25.1 & 7.2 & 4.8 & 78.3 & 9.0\\
IAP~\cite{lampert2014attribute} & 2.1 & 78.2 & 4.1 & 0.9 & 87.6 & 1.8 & 0.2 & {\textbf{72.8}} & 0.4 & 1.0 & 37.8 & 1.8 & 5.7 & 65.6 & 10.4\\
CMT~\cite{socher2013zero} & 8.4 & 86.9 & 15.3 & 8.7 & 89.0 & 15.9 & 4.7 & 60.1 & 8.7 & 8.7 & 28.0 & 13.3 & 10.9 & 74.2 & 19.0\\
DEVISE~\cite{frome2013devise} & 13.4 & 68.7 & 22.4 & 17.1 & 74.7 & 27.8 & 23.8 & 53.0 & 32.8 & 16.9 & 27.4 & 20.9 & 4.9 & 76.9 & 9.2\\
\hline
PSR~\cite{annadani2018preserving} & - & - & - & 20.7 & 73.8 & 32.3 & 24.6 & 54.3 & 33.9 & 20.8 & 37.2 & 26.7 & 13.5 & 51.4 & 21.4\\
DEM~\cite{zhang2017learning} & 32.8 & 84.7 & 47.3 & 30.5 & 86.4 & 45.1 & 19.6 & 57.9 & 29.2 & 20.5 & 34.3 & 25.6 & {\blue{11.1}} & 75.1 & {\blue{19.4}}\\ 
\hline
CONSE~\cite{norouzi2013zero} & 0.4 & 88.6 & 0.8 & 0.5 & 90.6 & 1.0 & 1.6 & {\blue{72.2}} & 3.1 & 6.8 & {\blue{39.9}} & 11.6 & 0.0 &  {\textbf{91.2}} & 0.0\\
SSE~\cite{zhang2015zero} & 7.0 & 80.5 & 12.9 & 8.1 & 82.5 & 14.8 & 8.5 & 46.9 & 14.4 & 2.1 & 36.4 & 4.0 & 0.2 & 78.9 & 0.4\\
LATEM~\cite{xian2016latent} & 7.3 & 71.7 & 13.3 & 11.5 & 77.3 & 20.0 & 15.2 & 57.3 & 24.0 & 14.7 & 28.8 & 19.5 & 0.1 & 73.0 & 0.2\\
ALE~\cite{akata2016label} & 16.8 & 76.1 & 27.5 & 14.0 & 81.8 & 23.9 & {{23.7}} & 62.8 & 34.4 & 21.8 & 33.1 & 26.3 & 4.6 & 73.7 & 8.7\\
SJE~\cite{akata2015evaluation} & 11.3 & 74.6 & 19.6 & 8.0 & 73.9 & 14.4 & 23.5 & 59.2 & 33.6 & 14.7 & 30.5 & 19.8 & 3.7 & 55.7 & 6.9\\
ESZSL~\cite{romera2015embarrassingly} & 6.6 & 75.6 & 12.1 & 5.9 & 77.8 & 11.0 & 12.6 & 63.8 & 21.0 & 11.0 & 27.9 & 15.8 & 2.4 & 70.1 & 4.6\\
SYNC~\cite{changpinyo2016synthesized} & 8.9 & 87.3 & 16.2 & 10.0 & 90.5 & 18.0 & 11.5 & 70.9 & 19.8 & 7.9 & {\textbf{43.3}} & 13.4 & 7.4 & 66.3 & 13.3\\
SAE~\cite{kodirov2017semantic} & 1.8 & 77.1 & 3.5 & 1.1 & 82.2 & 2.2 & 7.8 & 54.0 & 13.6 & 8.8 & 18.0 & 11.8 & 0.4 & 80.9 & 0.9\\
GFZSL~\cite{verma2017simple} & 1.8 & 80.3 & 3.5 & 2.5 & 80.1 & 4.8 & 0.0 & 45.7 & 0.0 & 0.0 & 39.6 & 0.0 & 0.0 & 83.3 & 0.0\\
RN~\cite{Sung_2018_CVPR} & 31.4 &  {\textbf{91.3}} & 46.7 & 30.0 &  {\textbf{93.4}} & 45.3 & 38.1 & 61.4 & {{47.0}} & - & - & - & - & - & -\\
QFSL*~\cite{song2018transductive} & - & - & - & {\blue{66.2}} & 93.1 & {\blue{77.4}} & {\textbf{74.9}} & 71.5 & {\textbf{73.2}} & {\blue{31.2}} & {\textbf{51.3}} & {\blue{38.8}} & - & - & -\\
\hline
Ours & {\blue{36.9}} & {\blue{90.6}} & {\blue{52.4}} & 35.2 & 93.0 & 51.1 & 21.0 & 66.0 & 31.9 & 22.1 & 35.6 & 27.3 & 7.8 & 75.3 & 14.1\\
Ours\_ex &  {\textbf{71.4}} & 90.1 &  {\textbf{79.7}} &  {\textbf{68.4}} & {\blue{93.2}} &  {\textbf{78.9}} &  {\blue{54.0}} & 62.9 &  {\blue{58.1}} & {\textbf{47.2}} & 38.5 & {\textbf{42.4}} &  {\textbf{29.8}} & 79.4 & {\textbf{43.3}}\\
\hline
\end{tabular}
\end{center}
\caption{Recognition accuracy on five benchmarks under the generalized ZSL setting. \textbf{ts} denotes the recognition accuracy on unseen classes, and \textbf{tr} denotes the recognition accuracy on seen classes. 
\textbf{H} denotes the harmonic mean~\cite{xian2017zero}. The best results are in blod, and the second best results are in blue. (QFSL*: using test samples with {\textbf{extra supervision}} for model training and the Resnet101 backbone is {\textbf{fine-tuned}}.)}
\label{table:performance2}
\vspace{-0.3cm}
\end{table*}

\subsection{Comparison in conventional ZSL}\label{subsec:conv}
In this part, we evaluate our method in conventional ZSL setting. Our method, as well as other comparing methods, follow the same evaluation setting in ~\cite{xian2017zero}. Table~\ref{table:con_ZSL} shows the comparing results. It can be seen that out method obviously outperforms competitors which directly fix either the semantic space or the visual space as the embedding space. For example, our method outperforms DEVISE~\cite{frome2013devise} by 15.9\% on AwA1 dataset. In the comparison with two recent state-of-the-arts, PSR~\cite{annadani2018preserving} and DEM~\cite{zhang2017learning}, our method surpasses PSR~\cite{annadani2018preserving} by 2.7\% on AwA2 dataset, and even outperforms DEM~\cite{zhang2017learning} by 22.1\% on the SUN dataset. This is because the unidirectional embedding schemes fail to model the separability between different classes, as shown in Figure~\ref{fig:embedding} (a)(d), while we can learn a latent intermediate embedding space and explicitly maximize the inter-class separability as Eq.~\eqref{eq:eq1}. Although many competitors also propose to learn an intermediate embedding space, our method still observes improvement, since most of them ignore the intra-class compactness and exhibit large intra-class variance, which hinders their generalization capacity in ZSL. For example, our method surpasses CONSE~\cite{norouzi2013zero} by 24.5\% on the AwA1 dataset. Even comparing with the state-of-the-art RN~\cite{Sung_2018_CVPR}, our method still outperforms it constantly with a clear margin.

In addition, we can find that our method with the pseudo labeling strategy can further improve the performance obviously. In particular, on the CUB dataset, {\texttt{Ours\_ex}} outperforms the most competitive method, \ie, PSR~\cite{annadani2018preserving}, by 11.8\%, while on the AwA1 dataset, the improvement over the most competitive method, \ie, DEM~\cite{zhang2017learning}, is even up to 16.9\%. Since most competitors only utilize the supervision information from training data of seen classes for model training, the learned embedding space often produces bias towards seen classes. In contrast, the proposed pseudo labeling strategy incorporates unlabeled testing samples into model training, which can progressively balance the embedding space between seen and unseen classes, thus being able to better distinguish visual samples from unseen classes.

\subsection{Comparison in generalized ZSL}\label{subsec:gen}
In this part, we evaluate our method in generalized ZSL setting. Our method, as well as other comparing methods, follow the same evaluation setting in~\cite{xian2017zero}. The results are shown in Table~\ref{table:performance2}. In this setting, unseen testing samples may come form both seen and unseen classes, which makes the ZSL problem more challenging. Limited by ignoring the intra-class compactness or the inter-class separability in constructing the embedding space, most competitors fail to generalize well to unseen classes. For example, on AwA1 dataset, DAP~\cite{lampert2014attribute}, SSE~\cite{zhang2015zero} and SAE~\cite{kodirov2017semantic} \etc , obtain recognition accuracy on unseen classes below 10\%. In contrast, our method generalizes better to unseen classes in most cases. For example, compared with recent state-of-the-arts, DEM~\cite{zhang2017learning} and RN~\cite{Sung_2018_CVPR}, {\texttt{Ours}} improves the recognition accuracy on unseen classes by 4.7\% and 5.2\% on AwA2 dataset.

As mentioned above, most of the competitors suffer more from the bias problem in this setting, but our method with pseudo labelling strategy (\ie, \texttt{Ours\_ex}) shows better performance in most cases and surpasses its counterpart without utilizing testing data (\ie, \texttt{Ours}) with an obvious margin. For example, compared with other competitors on AwA1 dataset, {\texttt{Ours\_ex}} improves the recognition accuracy on unseen classes by 39.6\% and above. QFSL~\cite{song2018transductive} is another work that utilizes testing samples in the training process, thus under \emph{transductive setting} as well. It obtains satisfactory performance. But their results are not directly comparable to ours. First, they use extra supervision information that they know a testing sample is from a seen or unseen class in advance. But we are blind to this information. Second, they fine-tune the ResNet101 backbone, which is frozen in our network training. 

\begin{figure}
\setlength{\abovecaptionskip}{0pt}
\begin{center}
\subfigure[ZSL performance]{\includegraphics[height=1.2in,width=1.6in,angle=0]{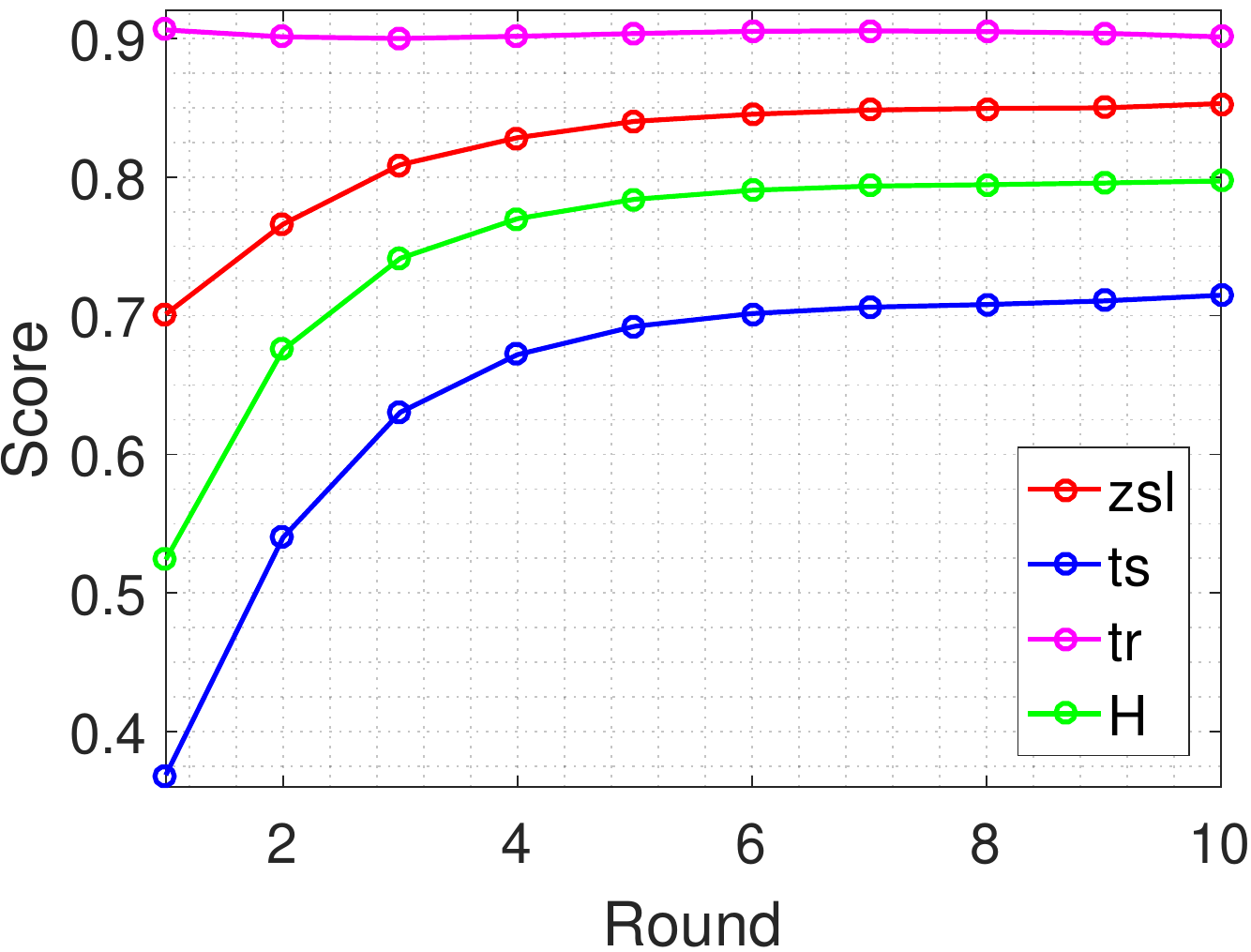}}
\subfigure[Accuracy of pseudo labels]{\includegraphics[height=1.2in,width=1.6in,angle=0]{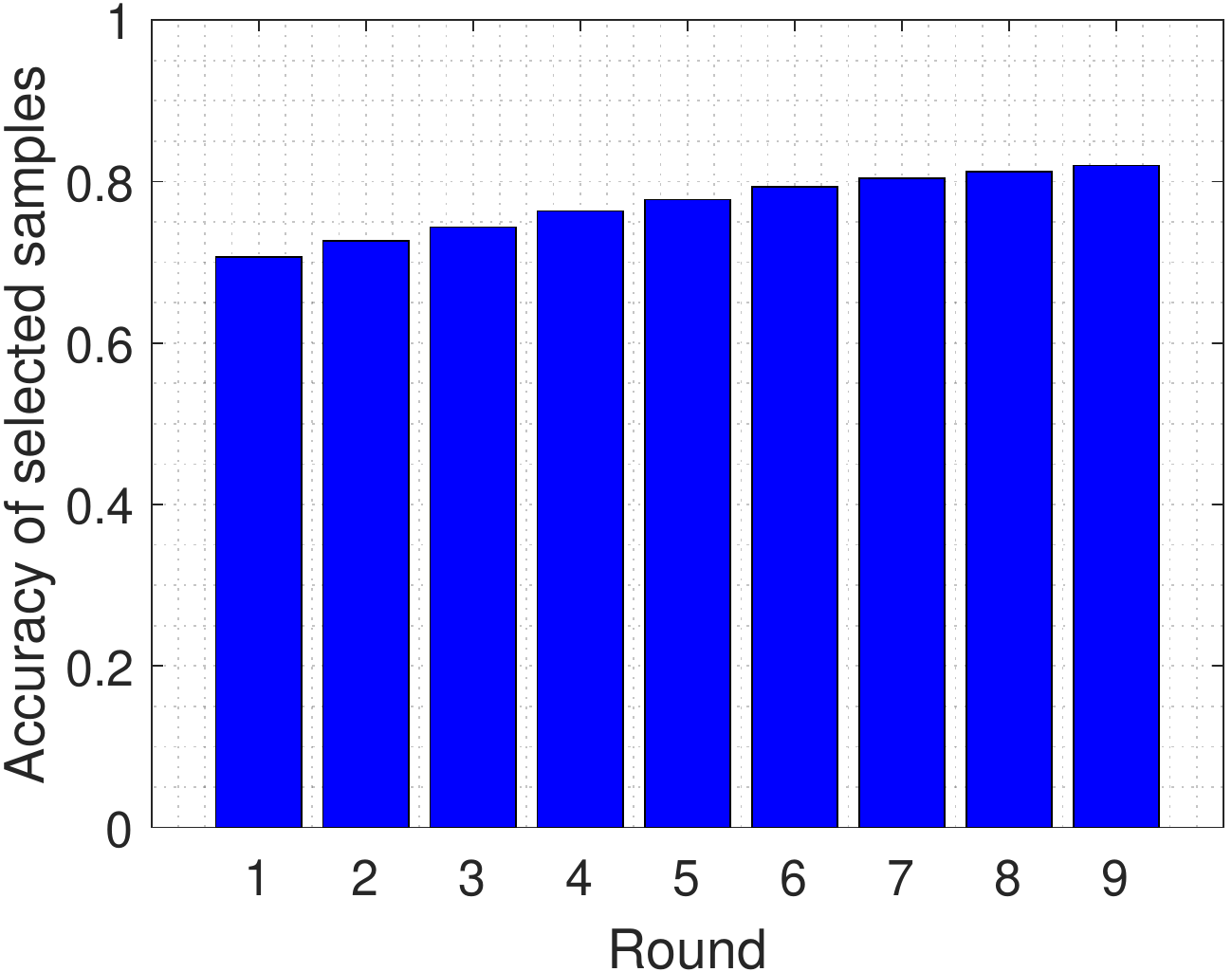}}
\end{center}
\vspace{-0.3cm}
\caption{Recognition accuracy of {\texttt{Ours\_ex}} and prediction accuracy of selected unseen class samples versus round (\eg, $R$) of model calibration in Algorithm~\ref{alg:Algorithm2} on AwA1 dataset. 'zsl' denotes the accuracy under conventional ZSL setting. 'ts', 'tr' and 'H' denotes measures (\eg, accuracy on unseen classes, accuracy on seen classes and the harmonic mean) under generalized ZSL setting.}
\vspace{-0.3cm}
\label{fig:round}
\end{figure}

\subsection{Further discussion}
In this subsection, we will conduct the ablation study for the proposed pseudo labeling strategy and the sensitivity analysis of parameter $\lambda$.

\paragraph{Ablation study for pseudo labeling strategy} To clarify this point, we plot the recognition accuracy curves of {\texttt{Ours\_ex}} on AwA1 dataset under both conventional and generalized ZSL settings versus the round $R$ of model calibration in Figure~\ref{fig:round} (a). It can be seen that with the increment of rounds, the recognition accuracy on unseen classes is gradually improved and ultimately converges. The reason is intuitive. At the beginning, to avoid introducing too many incorrectly labeled testing samples to mislead the calibration, we only select very limited testing samples from predicted unseen classes which fail to represent the comprehensive distribution of unseen classes. The learned embedding space thus can be further improved. With more rounds of calibration, more and more correctly labeled testing samples from unseen classes are introduced and the capacity of the embedding space in distinguishing unseen classes is gradually enhanced. It will ultimately converge when no new information is introduced. In contrast, the recognition accuracy on seen classes stays stable. This is because no testing samples are introduced from seen class. We also tried to select testing samples from seen classes for calibration. However, no obvious improvement on seen classes is observed, since extensive training samples from seen classes already contain sufficient information for discriminating seen classes, \eg, the recognition accuracy of {\texttt{Ours}} on seen classes is up to $90.6\%$ on AwA1 dataset in Table~\ref{table:performance2}.

The success of pseudo labeling strategy depends on the premise that most of the selected testing samples are labeled correctly. To clarify this point, we show the prediction accuracy of selected samples from unseen class in each round of model calibration under generalized ZSL setting in Figure~\ref{fig:round}(b). We find that the prediction accuracy in the first round is over 70\%. With the increase of round, the prediction accuracy can be further slightly increased.

\begin{figure}
\setlength{\abovecaptionskip}{0pt}
\begin{center}
\subfigure[Ours]{\includegraphics[height=1.2in,width=1.6in,angle=0]{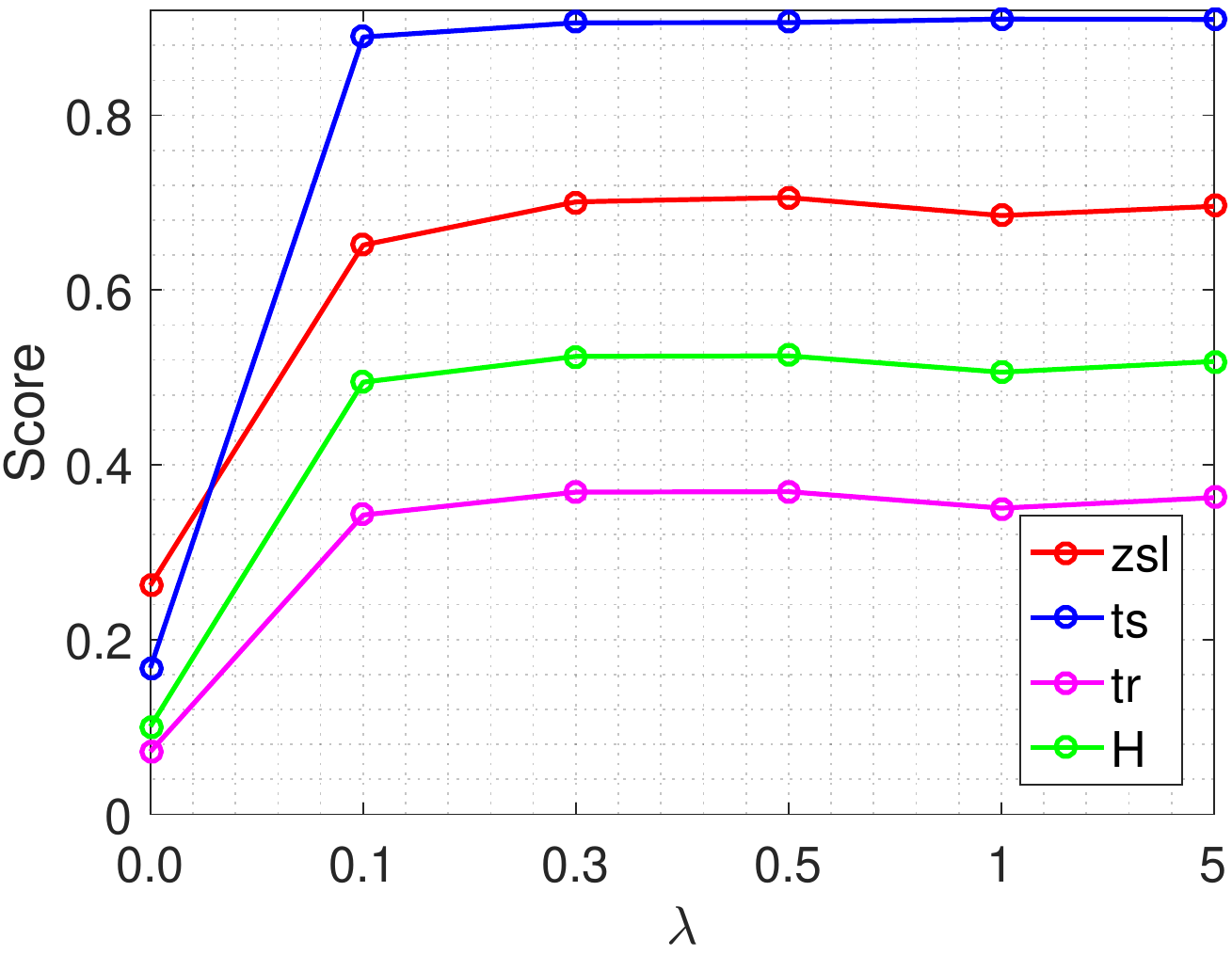}}
\subfigure[Ours\_ex]{\includegraphics[height=1.2in,width=1.6in,angle=0]{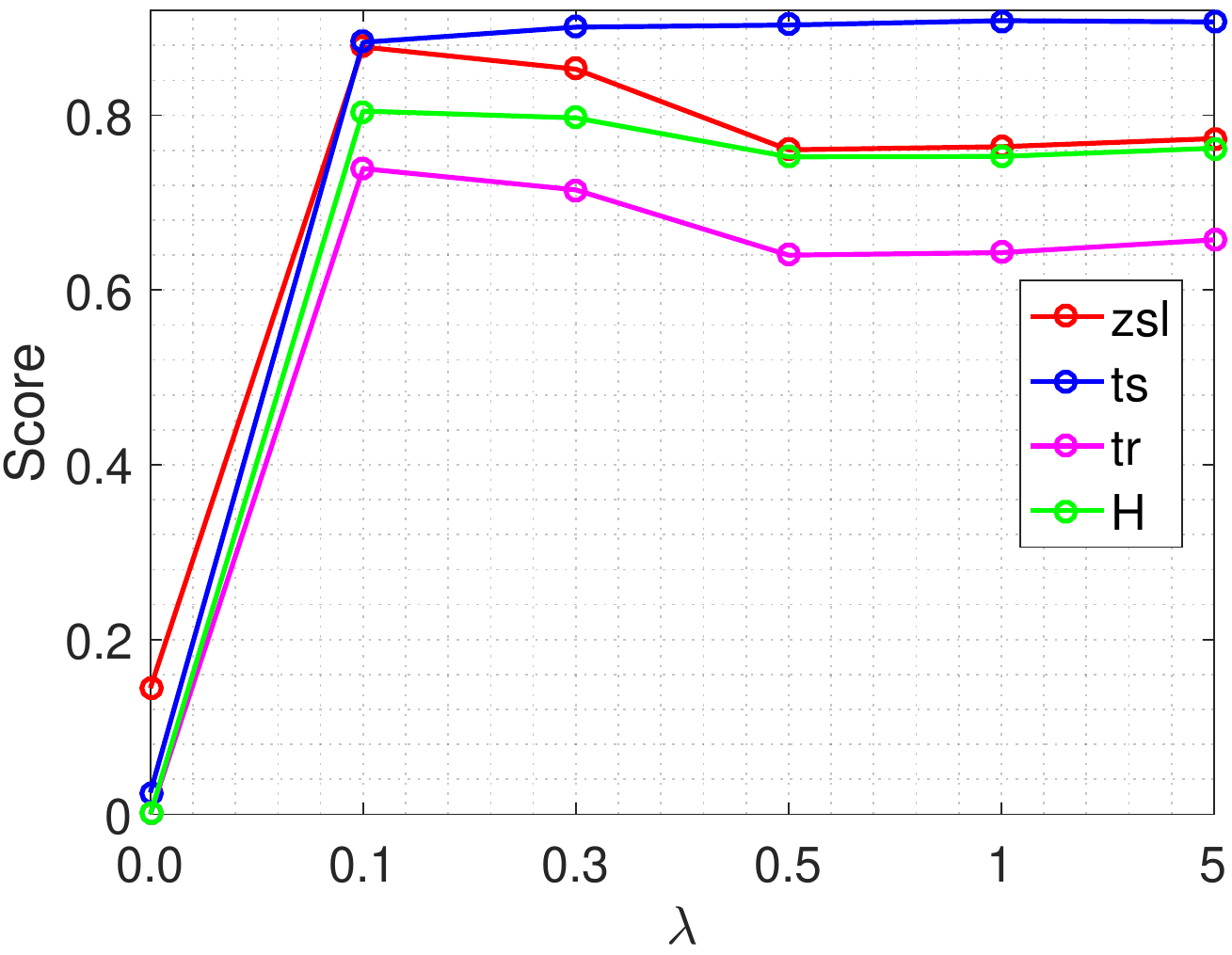}}
\end{center}
\vspace{-0.3cm}
\caption{Recognition accuracy of our methods on AwA1 dataset with different $\lambda$ under both conventional and generalized ZSL settings. 'zsl', 'ts', 'tr' and 'H' are the same as that in Figure~\ref{fig:round}.}
\vspace{-0.3cm}
\label{fig:lamb}
\end{figure}

\paragraph{Sensitivity analysis of $\lambda$} 
To demonstrate the effect of $\lambda$ to the performance of the proposed method, we evaluate {\texttt{Ours}} and {\texttt{Ours\_ex}} with different $\lambda$s on the AwA1 dataset under both conventional and generalized ZSL settings. The results are summarized into Figure~\ref{fig:lamb}. It can be seen that our methods perform stably within a wide range of $\lambda$, especially {\texttt{Ours}}. When $\lambda=0$, the resultant embedding model collapses obviously. This is owing to that we separately map the visual samples and the semantic descriptions into a latent embedding space. When $\lambda=0$, trivial solutions will be reached, \eg, visual samples and semantic descriptions from different classes are mapped into the same point or zeros. Therefore, regularizing the inter-class separability with a non-zero $\lambda$ is crucial for the proposed method.


\section{Conclusion}
In this study, we conducted an in-depth study on the construction of embedding space for ZSL and posited that an ideal embedding space should satisfy two criteria: intra-class compactness and inter-class separability. Towards this goal, we presented a simple but effective two-branch network to simultaneously map semantic descriptions and visual samples into a joint space, on which visual embeddings are forced to regress to their class-level semantic embeddings and the embeddings crossing classes are required to be distinguishable by a trainable classifier. In addition, we developed a pseudo labeling strategy to extend our method to the transductive setting to handle the model bias problem in ZSL. Experimental results demonstrated the effectiveness of the proposed method as well as its transductive extension.

{\small
\bibliographystyle{ieee}
\bibliography{egbib}

\begin{thebibliography}{10}\itemsep=-1pt

\bibitem{akata2016label}
Z.~Akata, F.~Perronnin, Z.~Harchaoui, and C.~Schmid.
\newblock Label-embedding for image classification.
\newblock {\em IEEE transactions on pattern analysis and machine intelligence},
  38(7):1425--1438, 2016.

\bibitem{akata2015evaluation}
Z.~Akata, S.~Reed, D.~Walter, H.~Lee, and B.~Schiele.
\newblock Evaluation of output embeddings for fine-grained image
  classification.
\newblock In {\em Proceedings of the IEEE Conference on Computer Vision and
  Pattern Recognition}, pages 2927--2936, 2015.

\bibitem{annadani2018preserving}
Y.~Annadani and S.~Biswas.
\newblock Preserving semantic relations for zero-shot learning.
\newblock In {\em Proceedings of the IEEE Conference on Computer Vision and
  Pattern Recognition}, pages 7603--7612, 2018.

\bibitem{changpinyo2016synthesized}
S.~Changpinyo, W.-L. Chao, B.~Gong, and F.~Sha.
\newblock Synthesized classifiers for zero-shot learning.
\newblock In {\em Proceedings of the IEEE Conference on Computer Vision and
  Pattern Recognition}, pages 5327--5336, 2016.

\bibitem{chao2016empirical}
W.-L. Chao, S.~Changpinyo, B.~Gong, and F.~Sha.
\newblock An empirical study and analysis of generalized zero-shot learning for
  object recognition in the wild.
\newblock In {\em European Conference on Computer Vision}, pages 52--68.
  Springer, 2016.

\bibitem{farhadi2009describing}
A.~Farhadi, I.~Endres, D.~Hoiem, and D.~Forsyth.
\newblock Describing objects by their attributes.
\newblock In {\em Computer Vision and Pattern Recognition, 2009. CVPR 2009.
  IEEE Conference on}, pages 1778--1785. IEEE, 2009.

\bibitem{frome2013devise}
A.~Frome, G.~S. Corrado, J.~Shlens, S.~Bengio, J.~Dean, T.~Mikolov, et~al.
\newblock Devise: A deep visual-semantic embedding model.
\newblock In {\em Advances in neural information processing systems}, pages
  2121--2129, 2013.

\bibitem{fu2016semi}
Y.~Fu and L.~Sigal.
\newblock Semi-supervised vocabulary-informed learning.
\newblock In {\em Proceedings of the IEEE Conference on Computer Vision and
  Pattern Recognition}, pages 5337--5346, 2016.

\bibitem{he2016deep}
K.~He, X.~Zhang, S.~Ren, and J.~Sun.
\newblock Deep residual learning for image recognition.
\newblock In {\em Proceedings of the IEEE conference on computer vision and
  pattern recognition}, pages 770--778, 2016.

\bibitem{kingma2014adam}
D.~P. Kingma and J.~Ba.
\newblock Adam: A method for stochastic optimization.
\newblock {\em arXiv preprint arXiv:1412.6980}, 2014.

\bibitem{kodirov2015unsupervised}
E.~Kodirov, T.~Xiang, Z.~Fu, and S.~Gong.
\newblock Unsupervised domain adaptation for zero-shot learning.
\newblock In {\em Proceedings of the IEEE International Conference on Computer
  Vision}, pages 2452--2460, 2015.

\bibitem{kodirov2017semantic}
E.~Kodirov, T.~Xiang, and S.~Gong.
\newblock Semantic autoencoder for zero-shot learning.
\newblock In {\em 2017 IEEE Conference on Computer Vision and Pattern
  Recognition (CVPR)}, pages 4447--4456. IEEE, 2017.

\bibitem{lampert2014attribute}
C.~H. Lampert, H.~Nickisch, and S.~Harmeling.
\newblock Attribute-based classification for zero-shot visual object
  categorization.
\newblock {\em IEEE Transactions on Pattern Analysis and Machine Intelligence},
  36(3):453--465, 2014.

\bibitem{lecun2015deep}
Y.~LeCun, Y.~Bengio, and G.~Hinton.
\newblock Deep learning.
\newblock {\em nature}, 521(7553):436, 2015.

\bibitem{lei2015predicting}
J.~Lei~Ba, K.~Swersky, S.~Fidler, et~al.
\newblock Predicting deep zero-shot convolutional neural networks using textual
  descriptions.
\newblock In {\em Proceedings of the IEEE International Conference on Computer
  Vision}, pages 4247--4255, 2015.

\bibitem{liang2015recurrent}
M.~Liang and X.~Hu.
\newblock Recurrent convolutional neural network for object recognition.
\newblock In {\em Proceedings of the IEEE Conference on Computer Vision and
  Pattern Recognition}, pages 3367--3375, 2015.

\bibitem{norouzi2013zero}
M.~Norouzi, T.~Mikolov, S.~Bengio, Y.~Singer, J.~Shlens, A.~Frome, G.~S.
  Corrado, and J.~Dean.
\newblock Zero-shot learning by convex combination of semantic embeddings.
\newblock {\em arXiv preprint arXiv:1312.5650}, 2013.

\bibitem{patterson2012sun}
G.~Patterson and J.~Hays.
\newblock Sun attribute database: Discovering, annotating, and recognizing
  scene attributes.
\newblock In {\em Computer Vision and Pattern Recognition (CVPR), 2012 IEEE
  Conference on}, pages 2751--2758. IEEE, 2012.

\bibitem{qiao2016less}
R.~Qiao, L.~Liu, C.~Shen, and A.~van~den Hengel.
\newblock Less is more: zero-shot learning from online textual documents with
  noise suppression.
\newblock In {\em Proceedings of the IEEE Conference on Computer Vision and
  Pattern Recognition}, pages 2249--2257, 2016.

\bibitem{romera2015embarrassingly}
B.~Romera-Paredes and P.~Torr.
\newblock An embarrassingly simple approach to zero-shot learning.
\newblock In {\em International Conference on Machine Learning}, pages
  2152--2161, 2015.

\bibitem{santoro2017simple}
A.~Santoro, D.~Raposo, D.~G. Barrett, M.~Malinowski, R.~Pascanu, P.~Battaglia,
  and T.~Lillicrap.
\newblock A simple neural network module for relational reasoning.
\newblock In {\em Advances in neural information processing systems}, pages
  4967--4976, 2017.

\bibitem{shigeto2015ridge}
Y.~Shigeto, I.~Suzuki, K.~Hara, M.~Shimbo, and Y.~Matsumoto.
\newblock Ridge regression, hubness, and zero-shot learning.
\newblock In {\em Joint European Conference on Machine Learning and Knowledge
  Discovery in Databases}, pages 135--151. Springer, 2015.

\bibitem{socher2013zero}
R.~Socher, M.~Ganjoo, C.~D. Manning, and A.~Ng.
\newblock Zero-shot learning through cross-modal transfer.
\newblock In {\em Advances in neural information processing systems}, pages
  935--943, 2013.

\bibitem{song2018transductive}
J.~Song, C.~Shen, Y.~Yang, Y.~Liu, and M.~Song.
\newblock Transductive unbiased embedding for zero-shot learning.
\newblock In {\em Proceedings of the IEEE Conference on Computer Vision and
  Pattern Recognition}, pages 1024--1033, 2018.

\bibitem{Sung_2018_CVPR}
F.~Sung, Y.~Yang, L.~Zhang, T.~Xiang, P.~H. Torr, and T.~M. Hospedales.
\newblock Learning to compare: Relation network for few-shot learning.
\newblock In {\em The IEEE Conference on Computer Vision and Pattern
  Recognition (CVPR)}, June 2018.

\bibitem{verma2017simple}
V.~K. Verma and P.~Rai.
\newblock A simple exponential family framework for zero-shot learning.
\newblock In {\em Joint European Conference on Machine Learning and Knowledge
  Discovery in Databases}, pages 792--808. Springer, 2017.

\bibitem{vinyals2016matching}
O.~Vinyals, C.~Blundell, T.~Lillicrap, D.~Wierstra, et~al.
\newblock Matching networks for one shot learning.
\newblock In {\em Advances in Neural Information Processing Systems}, pages
  3630--3638, 2016.

\bibitem{wah2011multiclass}
C.~Wah, S.~Branson, P.~Perona, and S.~Belongie.
\newblock Multiclass recognition and part localization with humans in the loop.
\newblock In {\em Computer Vision (ICCV), 2011 IEEE International Conference
  on}, pages 2524--2531. IEEE, 2011.

\bibitem{xian2016latent}
Y.~Xian, Z.~Akata, G.~Sharma, Q.~Nguyen, M.~Hein, and B.~Schiele.
\newblock Latent embeddings for zero-shot classification.
\newblock In {\em Proceedings of the IEEE Conference on Computer Vision and
  Pattern Recognition}, pages 69--77, 2016.

\bibitem{xian2017zero}
Y.~Xian, C.~H. Lampert, B.~Schiele, and Z.~Akata.
\newblock Zero-shot learning-a comprehensive evaluation of the good, the bad
  and the ugly.
\newblock {\em IEEE transactions on pattern analysis and machine intelligence},
  2018.

\bibitem{zhang2018unsupervised}
L.~Zhang, P.~Wang, W.~Wei, H.~Lu, C.~Shen, A.~van~den Hengel, and Y.~Zhang.
\newblock Unsupervised domain adaptation using robust class-wise matching.
\newblock {\em IEEE Transactions on Circuits and Systems for Video Technology},
  2018.

\bibitem{zhang2017learning}
L.~Zhang, T.~Xiang, and S.~Gong.
\newblock Learning a deep embedding model for zero-shot learning.
\newblock In {\em Proceedings of the IEEE Conference on Computer Vision and
  Pattern Recognition}, pages 2021--2030, 2017.

\bibitem{zhang2015zero}
Z.~Zhang and V.~Saligrama.
\newblock Zero-shot learning via semantic similarity embedding.
\newblock In {\em Proceedings of the IEEE international conference on computer
  vision}, pages 4166--4174, 2015.

\bibitem{zhang2016zero}
Z.~Zhang and V.~Saligrama.
\newblock Zero-shot learning via joint latent similarity embedding.
\newblock In {\em Proceedings of the IEEE Conference on Computer Vision and
  Pattern Recognition}, pages 6034--6042, 2016.

\end{thebibliography}
}

\end{document}